\documentclass[10pt,twocolumn,letterpaper]{article}

\usepackage{cvpr}
\usepackage{times}
\usepackage{epsfig}
\usepackage{graphicx}
\usepackage{amsmath}
\usepackage{amssymb}

\usepackage[ruled]{algorithm2e}
\usepackage{multirow}
\usepackage{array}
\usepackage{ctable} 
\usepackage{mathptmx}
\usepackage{makecell}
\usepackage{arydshln}
\usepackage[capposition=bottom]{floatrow}
\usepackage{comment}
\DeclareMathAlphabet{\mathcal}{OMS}{cmsy}{m}{n}
\usepackage{textcomp}
\usepackage{dblfloatfix}

\newcolumntype{?}[1]{!{\vrule width #1}}

\usepackage[pagebackref=true,breaklinks=true,letterpaper=true,colorlinks=false,bookmarks=false]{hyperref}

\cvprfinalcopy 

\def\cvprPaperID{999} 

\ifcvprfinal\fi

\begin{document}

\title{Im2Flow: Motion Hallucination from Static Images for Action Recognition}
\author{Ruohan Gao\\
UT Austin\\
{\tt\small rhgao@cs.utexas.edu}
\and
Bo Xiong\\
UT Austin\\
{\tt\small bxiong@cs.utexas.edu}
\and
Kristen Grauman\\
UT Austin\\
{\tt\small grauman@cs.utexas.edu}
}

\maketitle

\begin{abstract}
Existing methods to recognize actions in static images take the images at their face value, 
learning the appearances---objects, scenes, and body poses---that distinguish each action class.
However, such models are deprived of the rich dynamic structure and motions that also define human activity. 
We propose an approach that hallucinates the unobserved future motion implied by a single snapshot to help static-image action recognition.
The key idea is to learn a prior over short-term dynamics from thousands of unlabeled videos, infer the anticipated optical flow on novel static images, and then train discriminative models that exploit both streams of information.
Our main contributions are twofold.  First, we devise an encoder-decoder convolutional neural network and a novel optical flow encoding that can translate a static image into an accurate flow map.  Second, we show the power of hallucinated flow for recognition, successfully transferring the learned motion into a standard two-stream network for activity recognition.  On seven datasets, we demonstrate the power of the approach.  It not only achieves state-of-the-art accuracy for dense optical flow prediction, but also consistently enhances recognition of actions and dynamic scenes.
\end{abstract}
\section{Introduction}
\label{sec:intro}

Video-based action recognition has long been an active research topic in computer vision~\cite{efros2003recognizing,jhuang2007biologically,laptev2008learning,wang2011action,wang2013action}, with many recent methods employing deep Convolutional Neural Networks (CNNs)~\cite{twostream,karpathy2014large,ji20133d,yue2015beyond,varol2016long,feichtenhofer2016convolutional,wang2017temporal}.
Regardless of the approach, most methods rely on two crucial and complementary cues: appearance and motion. Motion is usually represented by (stacked) optical flow or flow-based descriptors~\cite{twostream,feichtenhofer2016convolutional,efros2003recognizing,lin2009recognizing,wang2017temporal,Girdhar_17a_ActionVLAD}, localized spatio-temporal descriptors~\cite{laptev2003space,willems2008efficient} or trajectories~\cite{wang2011action,wang2013action}.

\begin{figure}
    \center
    \includegraphics[scale=0.95]{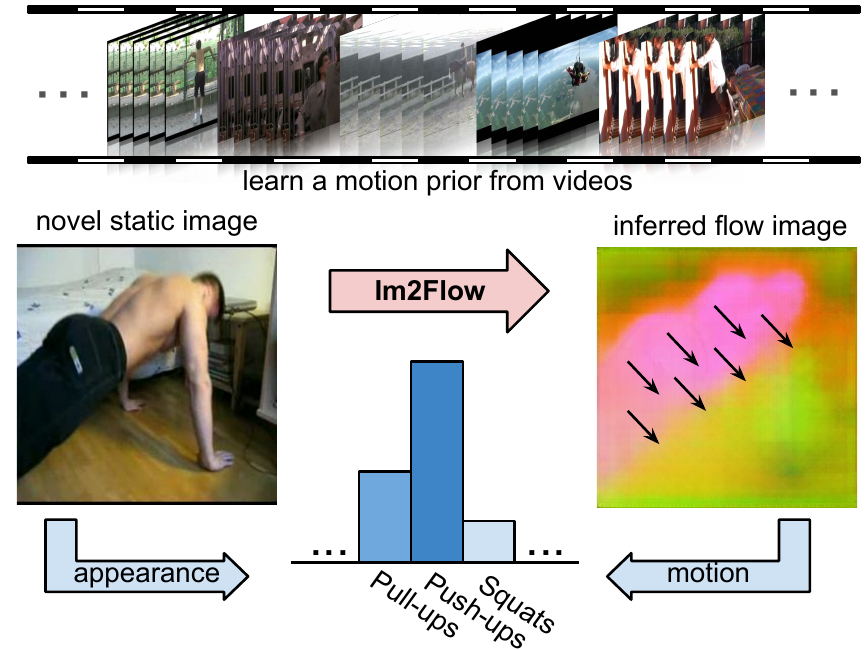}
    \vspace{-4pt}
    \caption{Our system first learns a motion prior by watching thousands of video clips containing various actions. Then, given a static image, the system translates the observed RGB image into a flow map encoding the inferred motion for the static image.  
    Finally, we combine the appearance of the original static image and the motion from the inferred flow to perform action recognition.}
    \label{fig:concept}
    \vspace{-12pt}
\end{figure}

\emph{Static-image action recognition} instead requires the system to identify the activity taking place in an individual photo~\cite{guo2014survey}.
The problem is of great practical interest for organizing photo collections (e.g., on the Web, social media, stock photos) based on human behavior and events. Yet, it presents the additional challenge of understanding activity in the absence of motion information.

Or is the motion really absent? A static snapshot can reveal the motions that are likely to occur next, at least for human viewers. Indeed, neuroscientists report that the medial temporal/medial superior temporal cortex---one of the main brain regions engaged in the perceptual analysis of visual motion---is also involved in representing implied motion from static images~\cite{kourtzi2000activation}. Over many years of observations, humans accumulate visual experience about how things move in the world. Given a single static image, not only can we interpret the instantaneous semantic content, but also we can anticipate what is going to happen next, e.g., based on human poses and object configurations present in the image.  This suggests that a human viewer can leverage the \emph{implied} motion to help perceive actions in static images. For example, given a static image as shown in Fig.~\ref{fig:concept}, expecting that the person's back is going to either move up or down may aid the recognition of push-ups.

We propose an approach for static-image action recognition that is inspired by this notion of visual dynamics accumulated from past temporal observations. The main idea is to acquire from videos a model for how objects and people move, then embed the resulting knowledge into a representation for individual images. In this way, even when limited to just one moment of observation (a single image), action recognition can be informed by the anticipated dynamics. In particular, we train a deep network to learn a motion prior from a large set of unlabeled videos, and then transfer the learned motion from videos to static images to hallucinate their motion. We devise  an effective encoding for optical flow (Sec.~\ref{sec:motion_encoding}) as well as an encoder-decoder network to learn the motion prior from videos (Sec.~\ref{sec:network}). Finally, we leverage the predicted motion to aid action recognition, by combining the appearance from the original static image and the motion from the inferred flow.  

On seven challenging datasets for recognition of actions and dynamic scenes, our approach yields a significant accuracy boost by incorporating the hallucinated motion. Importantly, we also demonstrate that our approach is beneficial even in the case where the motion prior training videos do not contain the same actions as the static images.

We make two main contributions. First, we formulate motion prediction as a novel image-to-image translation framework, and achieve state-of-the-art performance on dense optical flow prediction from static images, improving substantially on prior formulations for flow estimation~\cite{cmu-opticalflow-iccv2015,dejavu-eccv2014}. Secondly, we explore how implied motion aids action recognition. We show how injecting inferred motion into a standard two-stream network achieves significant gains compared to the one-stream counterpart, and we obtain state-of-the-art accuracy for multiple benchmarks.

\vspace*{-0.05in}

\section{Related Work}

Our work relates to action recognition, visual anticipation, and image-to-image translation.
\vspace*{-0.15in}

\paragraph{Action Recognition}
Video-based action recognition is a well-studied problem. Various video representations have been proposed to utilize both appearance and motion, including hand-crafted local spatio-temporal features~\cite{laptev2003space,willems2008efficient,wang2011action,wang2013action}; mid-level features~\cite{raptis2012discovering,jain2013representing,wang2013motionlets}; and deeply-learned features~\cite{twostream,karpathy2014large,yue2015beyond,feichtenhofer2016convolutional,varol2016long}. Recent work aims to model long term temporal structure, via recurrent neural networks~\cite{yue2015beyond,donahue2015long}, ranking functions~\cite{fernando2015modeling}, or pooling across space and time~\cite{Girdhar_17a_ActionVLAD}.

In static images, due to the absence of temporal information, various high-level cues are exploited, e.g., human body or body parts~\cite{thurau2008pose,maji2011action,yao2011human}, objects~\cite{prest2012weakly,yao2011human,sener2012recognizing}, human-object interactions~\cite{delaitre2011learning,chao2015hico}, and scene context~\cite{gupta2009observing,gkioxari2015contextual}. See~\cite{guo2014survey} for a comprehensive survey. Our work also targets action recognition in static images, but, unlike any of the above, we equip static images with dynamics learned from videos. To our knowledge, the only prior static-image approach to explicitly leverage video dynamics is~\cite{chen2013watching}. However, whereas~\cite{chen2013watching} leverages video to augment training images for the low-shot learning scenario, our method leverages video as a motion prior that enhances \emph{test} observations. Our experiments compare the two methods.

\vspace*{-0.15in}
\paragraph{Visual Anticipation}
Our work is also related to a wide body of work on visual future prediction~\cite{yuen2010data,kitani2012activity,walker2014patch,dejavu-eccv2014,fouhey2014predicting,cmu-opticalflow-iccv2015,walker2016uncertain,vondrick2016anticipating}. Most closely related are methods to predict optical flow from an image~\cite{walker2014patch,cmu-opticalflow-iccv2015,dejavu-eccv2014}. As we show in results, our formulation offers more accurate predictions. Furthermore, unlike any prior flow prediction work, we propose to integrate implied motion learned from thousands of unlabeled videos with action recognition. Note that while flow and depth are closely related problems, methods like DeepStereo~\cite{deepstereo} or DeepMorphing~\cite{deepmorph} assume two viewpoints as input to predict intermediate views. Other prediction tasks consider motion trajectories~\cite{walker2016uncertain,lee2017desire} and human body poses~\cite{fragkiadaki2015recurrent,chao2017forecasting}.  Another growing line of work aims to predict future frames in video~\cite{ranzato2014video,srivastava2015unsupervised,mathieu2016deep,xue2016visual,finn2016unsupervised,vondrick2016generating,vondrick2017generating,villegas2017learning}. Their goal is to generate images of good visual quality to illustrate the ``plausible futures", or potentially improve representation learning~\cite{srivastava2015unsupervised,finn2016unsupervised}.

In general, this line of work treats prediction as the end goal: e.g., predicting optical flow~\cite{cmu-opticalflow-iccv2015}; people's trajectories in a parking lot~\cite{kitani2012activity}; car movements on streets~\cite{walker2014patch}; or subsequent high-level events~\cite{yuen2010data,vondrick2016anticipating}.  In contrast, our objective is to infer motion as an auxiliary cue for action recognition in static images. Our idea bridges static-image recognition with video-level action understanding by transferring a motion prior learned from videos to images.

\vspace*{-0.15in}
\paragraph{Image-to-Image Translation}
Our technical solution for flow inference relates broadly to prior efforts to map an input pixel matrix directly to an output matrix.
Early work in so-called ``image-to-image translation" can be traced back to image analogies~\cite{hertzmann2001image}, where a nonparametric texture model is generated from a single input-output training image pair.
Recent work uses generative adversarial models~\cite{goodfellow2014generative} to perform image-to-image translation, with impressive results~\cite{pix2pix2017,liu2017unsupervised,CycleGAN2017}. Our dense optical flow prediction approach can be seen as a distinct image-to-image translation problem, where we encode the output motion space as a single ``image" matrix.  

\vspace*{-0.05in}

\section{Approach}
Our goal is to learn a motion prior from videos, and then transfer the motion prior to novel static images to enhance recognition. We first discuss how we encode optical flow for more reliable prediction (Sec.~\ref{sec:motion_encoding}). Then we present our Im2Flow network for motion prediction (Sec.~\ref{sec:network}). Finally, we describe how we use the predicted motion to help static-image action recognition (Sec.~\ref{sec:two-stream}).

\subsection{Motion Encoding}
\label{sec:motion_encoding}

Optical flow encodes the pattern of apparent motion of objects in a visual scene, and it is the most direct and common motion information used for action recognition. (Stacked) optical flow is frequently used as input to deep methods~\cite{twostream,feichtenhofer2016convolutional,varol2016long,wang2017temporal}. Often optical flow is represented by two separate grayscale images (matrices) that encode the quantized horizontal and vertical displacements.
However, since state-of-the-art pre-trained deep networks~\cite{alexnet,vgg,resnet} take a 3-channel RGB image as input, alternative encodings augment the two displacement maps with a third channel containing either the flow magnitude~\cite{donahue2015long} or all zeros~\cite{yue2015beyond}. In such encodings, the third channel stores either redundant or useless information.  Another approach is to encode the flow as an RGB image designed for visualization~\cite{baker2011database,fusionseg}, but this mapping is not reversible to obtain the motion vectors.  Prior work quantizes flow vectors to 40 clusters in order to treat flow estimation as classification~\cite{cmu-opticalflow-iccv2015}, but this has the drawback of providing only coarse flow estimates insufficient for recognition.

Our preliminary tests with the existing encodings confirmed these limitations (see~Supp.), leading us to develop a new encoding scheme well-suited to motion prediction via regression. Directly predicting the optical flow $(u,v)$ vector for each pixel in a static image is a highly under-constrained problem, and we hypothesize that detangling flow direction and strength may present an easier learning task. Therefore, we decouple the optical flow into the motion angle $\theta\in [0,2\pi]$ and magnitude $\mathcal{M}$. As we will describe in Sec.~\ref{sec:network}, we formulate flow prediction as a pixel-wise regression problem. Hence, direct prediction of $\theta$ is inappropriate because the angle in the coordinate system is circular (e.g., $2\pi$ is the same as 0). Therefore, we further divide $\theta$ into a horizontal direction and a vertical direction, represented by $\cos(\theta)$ and $\sin(\theta)$, respectively.  We encode optical flow as a single 3-channel flow image $\mathcal{F}$:
\begin{equation}
\vspace{-0.02in}
	\mathcal{F}_1 = \sin(\theta) = \frac{v}{\mathcal{M}};~~~~\mathcal{F}_2 = \cos(\theta) = \frac{u}{\mathcal{M}};~~~~\mathcal{F}_3 = \mathcal{M}.
\vspace{-0.02in}
\end{equation}
where $\mathcal{F}_i$ denotes the $i$-th channel.

Our motion encoding scheme has the following benefits: 1) It disentangles the convolved $(u,v)$ vector into three separate components, each indicating one important factor of motion, namely vertical direction, horizontal direction, and motion magnitude. This makes the high-dimensional motion prediction problem more factored; 2) It makes the regression of angle feasible, because $\sin(\theta)$ and $\cos(\theta)$ are non-circular and lie in the range of $[-1,1]$; 3) Encoding motion as a 3-channel image makes its usage efficient, convenient, and suitable for our framework, defined next.

\subsection{Im2Flow Network}
\label{sec:network}

\begin{figure*}[t]
    \center
    \includegraphics[scale=1.05]{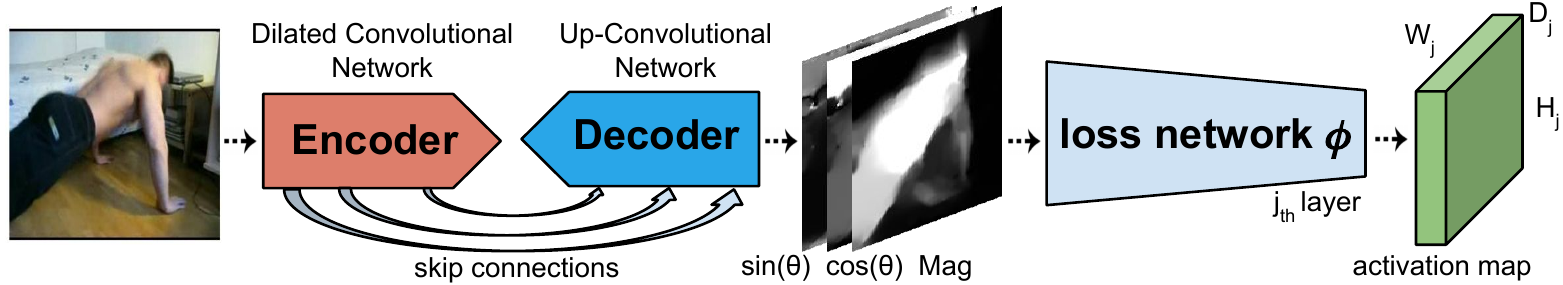}
    \caption{The network architecture of our Im2Flow framework. The network is an encoder-decoder that takes a static image as input and generates the corresponding 3-channel flow map $\mathcal{F}$ as output. Our training objective is a combination of the $L_2$ loss in the pixel space and in the deep feature space. A motion content loss network encourages the predicted flow image to preserve high-level motion features.}
    \label{fig:network}
    \vspace{-0.1in}
\end{figure*}

Let $X$ be the domain of static images containing an action, and let $Y$ consist of the corresponding flow images encoded as defined above. Our goal is to learn a mapping $G : X \rightarrow Y$ that will infer flow from an individual image.  During training, we are given ``labeled" pairs $\{x_i, y_i\}_{i=1}^{N}$ consisting of video frames $x_i$ and the true flow maps $y_i$ computed from surrounding frame(s) in the source video. We use the optical flow algorithm of~\cite{liu2009beyond} to automatically generate $y$ for training data, since it offers a good balance between speed and accuracy.\footnote{Accurate estimation of optical flow from real-world videos is a challenging problem on its own and is intensively studied in the literature~\cite{revaud2015epicflow,fischer2015flownet,kroeger2016fast,IMKDB17}. More accurate optical flow estimation could further improve the Im2Flow framework.} To mitigate the effects of noisy flow estimates stemming from realistic training videos (i.e., we train with YouTube data, cf.~Sec.~\ref{sec:expts}), following~\cite{cmu-opticalflow-iccv2015}, we average the optical flow of five future frames for each training image $x_i$ to compute the target $y_i$.

We devise a convolutional neural network (CNN) called Im2Flow to obtain $G$. Our Im2Flow network is an encoder-decoder, which takes a static image as input and outputs the predicted flow image $\hat{y} = G(x) = \mathcal{F}$. We adapt the U-Net architecture from~\cite{pix2pix2017} with some modifications, as illustrated in~Fig.~\ref{fig:network}. Both the encoder and decoder use modules of the form Convolution-BatchNorm-ReLU~\cite{ioffe2015batch}. We use dilated convolutions~\cite{yu2016multi} in the encoder. Dilated convolutions exponentially increase their receptive field size and maintain spatial resolution, which can capture long-range spatial dependencies. The decoder is an up-convolutional network that generates the predicted flow image. Skip connections connect the encoder and decoder to directly shuffle low-level information across the network, which is important for our dense motion prediction problem. See Supp. for the details of the complete architecture.  

Our Im2Flow network minimizes the combination of two losses: a pixel error loss and a motion content loss:
\vspace{-6.5pt}
\begin{equation}
\vspace{-4pt}
	L = L_{pixel} + \lambda L_{content}^{\phi,j}.
\vspace{-6pt}
\label{equ:loss}
\end{equation}
The pixel loss measures the agreement with the true flow:
\vspace{-2pt}
\begin{equation}
\vspace{-1pt}
	L_{pixel} = \mathbb{E}_{p,q\in \{x_i, y_i\}_{i=1}^{N}}\left[||y_i - G(x_i)||_2\right]	
\vspace{-5pt}
\end{equation}
for all pixels $p,q$ in the training images. It requires the training flow vectors to be accurately recovered.

The motion content loss enforces that the predicted motion image preserve high level motion features. It follows the spirit of previous perceptual loss functions~\cite{johnson2016perceptual}, but here for the sake of regularizing to realistic motion patterns. To represent realistic motion, we fine-tune an 18-layer ResNet~\cite{resnet} (pre-trained on ImageNet) for action classification on the UCF-101 dataset~\cite{soomro2012ucf101} using flow images as input. The motion content loss network $\phi$ is the resulting fine-tuned network. Then, we compute the $L_2$ loss on the activation maps extracted from the loss network $\phi$ for the predicted flow and ground-truth flow images. Hence, apart from encouraging the pixels of the output flow $G(x)$ to exactly match the pixels of the target flow $y$, we also encourage them to have similar high-level motion representations as computed by the loss network.

Note that our approach can operate with both unlabeled and labeled video. The supervision from action labels used in the motion content loss network slightly helps to transfer motion from videos to images, and enhances the results on static-image action recognition. As we show in the ablation study in Supp., our approach maintains substantial gains even if we completely remove supervision in our framework, i.e., our approach can learn solely from unlabeled video. Moreover, even if we do learn from labeled video, the test data is not assured to be from the same actions (cf.~Sec.~\ref{sec:action_recognition}).

Let $\phi_{j}(x)$ be the activations of the $j$-th layer of the network $\phi$ when processing the image $x$, and suppose $\phi_{j}(x)$ has a feature map of shape $D_j \times H_j \times W_j$.  The motion content loss is the (squared, normalized) Euclidean distance between the feature representations:
\begin{equation}
\vspace*{-2pt}
		L_{content}^{{\phi,j}} = \frac{1}{D_j \times H_j \times W_j} \mathbb{E}_{p,q\in \{x_i, y_i\}_{i=1}^{N}} \left[||\phi_j(y_i) - \phi_j(G(x_i))||_2\right].
\vspace*{-2pt}	
\end{equation}

We adjust optimization of the network to account for structure in our problem.  The angles for pixels of very low motion strength are less crucial, because these pixels tend to correspond to static scenes in the image, e.g., the bed and wall in the input image in Fig.~\ref{fig:network}. Therefore, the motion directions of these pixels are not as meaningful, and they usually originate from the camera motion. To require the network to focus more on predicting directions of the pixels that actually move, we weight the gradients of the first two channels by their motion magnitude. The network uses the weighted gradients to perform back-propagation. The weighting process forces the network to emphasize predictions on moving pixels.

The above architecture was the most effective among other alternatives we explored.  In particular, in preliminary experiments, we implemented a conditional generative adversary network (cGAN) to perform flow prediction, inspired by~\cite{pix2pix2017,CycleGAN2017}. In principle, such a GAN might handle multiple modes of motion ambiguity and predict flow images that encode realistic motion. However, we found that cGAN only helped to generate motion images of similar \emph{visual style} to the ground-truth flow maps: the color patterns were similar to the ground-truth, but the encoded motion was less accurate. Because the cGAN discriminator cares about differentiating real and fake outputs, the approach seems better suited to problems requiring output images of good visual quality, as opposed to our task, which requires precise pixel-wise estimates.

\vspace{-0.05in}
\subsection{Action Recognition with Implied Motion}
\label{sec:two-stream}

Recall our goal is twofold: to produce accurate flow maps for static images, and to explore their utility as auxiliary input for static image action recognition.  For the latter, we adopt the popular and effective two-stream CNN architecture~\cite{twostream} that is now widely used for CNN-based action recognition with videos~\cite{cheron2015p,donahue2015long,gkioxari2015finding,yue2015beyond,srivastava2015unsupervised,feichtenhofer2016convolutional,wang2017temporal,Girdhar_17a_ActionVLAD}.
The two-stream approach is designed to mimic the pathways of the human visual cortex for object recognition and motion recognition~\cite{goodale1992separate}.  Namely, the method decomposes video into spatial (RGB frames) and temporal (optical flow) components. These two components are fed into two separate deep networks. Each stream performs action recognition on its own and final predictions are computed as an average of the two outputs.

Along with being highly successful in the video literature, the two-stream approach is a natural fit for injecting our Im2Flow predictions into action recognition with static images. In short, we augment both training and testing images with their respective inferred flow maps, then train the action recognition network with standard procedures.  See Sec.~\ref{sec:expts} for more details.

Why should the inferred motion help action learning with static images?  We hypothesize two rationales.  First, there is  value in elucidating a salient signal for action that can be difficult to learn directly from the images alone. 
Static images of different action classes can be visually similar, e.g., pull-ups vs.~push-ups, brushing teeth vs.~applying lipstick.  The motion implied for people and objects in the static images can help better distinguish subtle differences among such actions.  
 This parallels what is currently observed in the literature with real optical flow: presenting an action recognition network with explicit optical flow maps is much stronger than simply presenting the two source frames---even though the optical flow engine receives those same two frames~\cite{twostream,feichtenhofer2016convolutional,wang2017temporal}.  There is value in directing the learner's attention to a complex, high-dimensional signal that is useful but would likely require many orders of magnitude more data to learn simultaneously with the target recognition task.  Second, the Im2Flow network leverages a large amount of video to build a motion prior that regularizes the eventual activity learning process. Since action recognition datasets are relatively small w.r.t. the variability with which actions can be performed, a learning algorithm can easily overfit, e.g., to the background of training examples. The domain of inferred motion helps to get rid of elements irrelevant to the action performed in the image, and therefore mitigates overfitting.

\begin{table*}
\centering
\begin{tabular}{c?{0.5mm}ccc?{0.5mm}ccc?{0.5mm}ccc}
 \textbf{UCF-101} & EPE~$\downarrow$ & EPE-Canny & EPE-FG & DS~$\uparrow$ & DS-Canny & DS-FG & OS~$\uparrow$ & OS-Canny & OS-FG \\ \specialrule{.12em}{.1em}{.1em}
Pintea~\etal~~\cite{dejavu-eccv2014}  &  2.401         &   2.699  &  3.233         &   -0.001     &  -0.002 &   -0.005       &  0.513     &  0.544  &       0.555      \\ 
Walker~\etal~\cite{cmu-opticalflow-iccv2015}           &    2.391 &           2.696 &   3.139     &  0.003  &    0.001      &   0.014    &    0.661 &    0.673      &   0.662    \\ 
Nearest Neighbor   &  3.123       &  3.234   &    3.998       &   -0.002     & -0.001   &  -0.023        &   0.652    &   0.651  &   0.659              \\ 
Ours     &   \textbf{2.210}      &  \textbf{2.533}   &    \textbf{2.936}       &    \textbf{0.143}    & \textbf{0.135}   &    \textbf{0.137}             & \textbf{0.699}   &     \textbf{0.692}    &  \textbf{0.696}     \\ 
\specialrule{.12em}{.1em}{.1em}
\end{tabular}

\caption{Quantitative results of dense optical flow prediction on UCF-101. $\downarrow$ lower better, $\uparrow$ higher better. Across all measures, our method outperforms all baseline methods by a large margin. See Supp.~for similar results on HMDB-51 and Weizmann datasets.}
\label{table:optical_flow_eval}
\vspace*{-4pt}
\end{table*}

\begin{figure*}[t]
    \center
    \begin{minipage}{1\textwidth} 
    \includegraphics[scale=1.7]{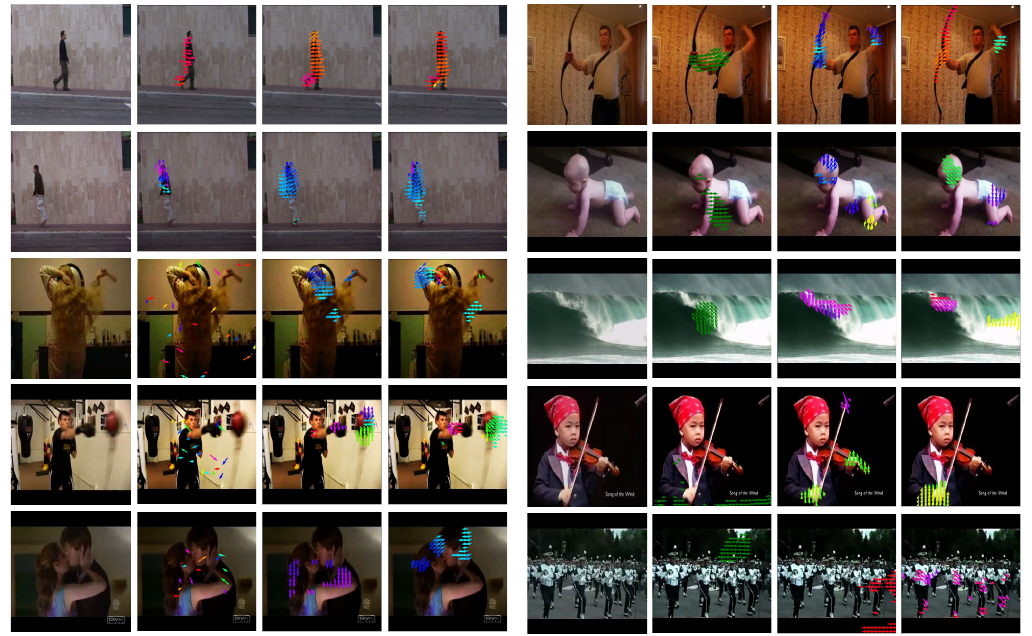}
    {\footnotesize \indent ~~~~(a) Input Image~~~~(b) Pintea~\etal~\cite{dejavu-eccv2014}~~~~~~~~~(c) Ours~~~~~~~~~~~(d) Ground-truth~~~~~~~~(a) Input Image~~~~~(b) Walker~\etal~\cite{cmu-opticalflow-iccv2015}~~~~~~~(c) Ours~~~~~~~~~~(d) Ground-truth}
    \vspace{0.1in}
\end{minipage}
\floatbox[{\capbeside\thisfloatsetup{capbesideposition={left,top},capbesidewidth=14cm}}]{figure}[\FBwidth]
{\caption{Examples of dense optical flow prediction (best viewed in color). The Pintea~\etal~\cite{dejavu-eccv2014} approach makes reasonable predictions on Weizmann (top left), but suffers on more complex datasets like HMDB-51 (bottom left). The Walker~\etal~\cite{cmu-opticalflow-iccv2015} approach often captures the general trend of motion, but the predicted motion is coarse. Our Im2Flow network accurately predicts motion that is much more fine-grained in various contexts. The last row shows two failure cases.  We use the color coding on the right for flow visualization.}\label{fig:qualitative}
}
{\includegraphics[width=1.8cm]{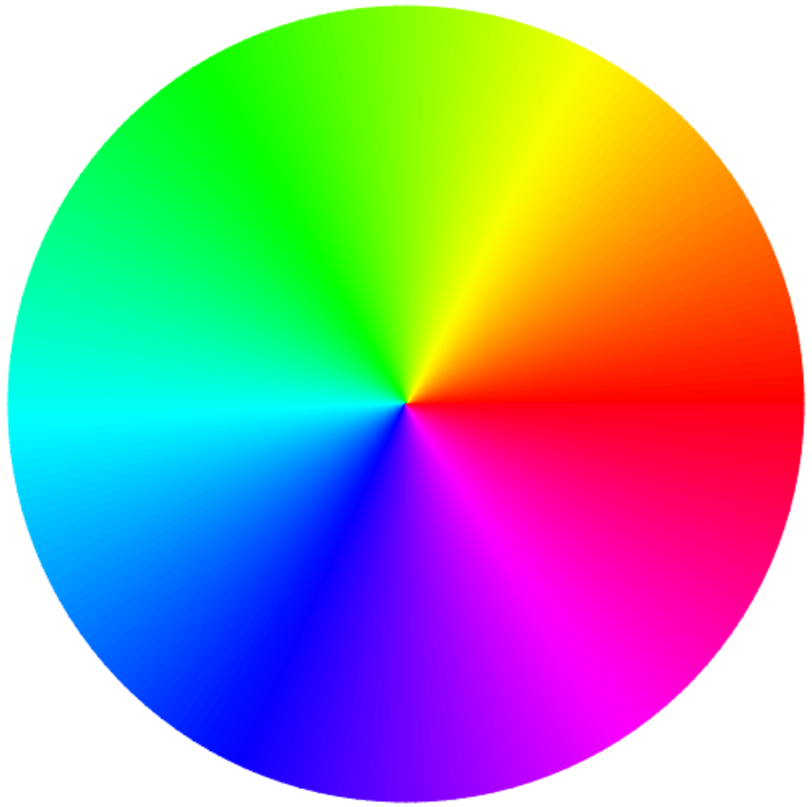}}
\vspace{-0.1in}
\end{figure*}

\vspace*{-4pt}
\section{Experiments}\label{sec:expts}

Using a total of 7 datasets, we validate our approach for 1) flow prediction accuracy (Sec.~\ref{sec:floweval}) and 2) action recognition from static images (Sec.~\ref{sec:action_recognition}
\vspace*{-0.2in}
\paragraph{Implementation details} We implement our Im2Flow network in Torch and train it with videos from UCF-101~\cite{soomro2012ucf101} and HMDB-51~\cite{kuehne2011hmdb}. We sample 500,000 frames from UCF-101 and 200,000 frames from HMDB-51 as our training data. We use minibatch SGD with a batch size of 32 and apply Adam solver~\cite{adamsolver}. We train with random horizontal flips and randomly cropped windows as data augmentation. For the motion content loss network, we use the activation maps after the second residual block of ResNet18 and we set $\lambda = 0.02$ in Equation~\eqref{equ:loss}. All our action recognition experiments are implemented in Caffe~\cite{jia2014caffe}. For action recognition, we use AlexNet~\cite{alexnet} with batch normalization~\cite{ioffe2015batch} as the base architecture for each stream.  We fuse the two streams' softmax prediction scores using the optimal weight determined on a validation set.

\subsection{Flow Prediction}\label{sec:floweval}

First, we directly evaluate Im2Flow's dense optical flow prediction.  Here we use three datasets: UCF-101~\cite{soomro2012ucf101}, HMDB-51~\cite{kuehne2011hmdb}, and Weizmann~\cite{ActionsAsSpaceTimeShapes_pami07}. For UCF-101 and HMDB-51, we hold out 10 videos from each class as test data and the rest as candidate training data; for Weizmann, we hold out the frames of \emph{shahar} as the test set. We compare with the following methods:
\vspace{-0.1in}
\begin{itemize}
\itemsep0em
\item \textbf{Walker~\etal~\cite{cmu-opticalflow-iccv2015}}: Existing CNN-based method that classifies each region in the image to a quantized optical flow vector. We use their publicly available model\footnote{\url{https://github.com/puffin444/optical_flow_prediction}}, which is trained on the whole UCF-101 dataset. 
\item \textbf{Pintea~\etal~\cite{dejavu-eccv2014}}: Existing structured random forest regression approach.  We use their publicly available code\footnote{\url{https://github.com/SilviaLauraPintea/DejaVu}} and train a model with default parameters.
\item \textbf{Nearest Neighbor}: Baseline that uses the pool5 features from a pre-trained AlexNet to retrieve the nearest training image, then adopts the ground-truth flow of that image.  Its training pool consists of the same frames that train Im2Flow.  This baseline is inspired by the method of Yuen \& Torralba~\cite{yuen2010data}, which identifies likely future events using nearest neighbor.
\end{itemize}
\vspace{-0.1in}

\vspace*{-12pt}
\paragraph{Evaluation metrics} We convert Im2Flow's outputs back to dense optical flow to compare against the ``ground truth" flow, which is computed from video with~\cite{liu2009beyond}.  
We employ a suite of metrics, following prior work in this area~\cite{dejavu-eccv2014,cmu-opticalflow-iccv2015}: End-Point-Error (EPE), Direction Similarity (DS), and Orientation Similarity (OS) (see Supp.~for details).
Apart from evaluating over all the pixels in the whole image, we also evaluate over masks on the 1) Canny edges, which 
 approximates measuring the error of moving pixels in simple scenes~\cite{dejavu-eccv2014,cmu-opticalflow-iccv2015}, and 2) 
 foreground (FG) regions (computed with~\cite{pixelobjectness}), which
often correspond to the moving objects.

\begin{table*}
	\fontsize{8.5}{10.5} \selectfont
    \centering
    \begin{tabular}{@{}cc?{0.5mm}*{6}{c}}
    \multicolumn{1}{c}{} & &UCF-static    &HMDB-static    &PennAction    &Willow    &Stanford10  &PASCAL2012     \\ \specialrule{.12em}{.1em}{.1em}
 	& Appearance Stream    &63.6   &35.1   &73.1  &65.1  &81.3 & 65.0 \\ \specialrule{.12em}{.1em}{.1em}
 	\multirow{5}*{\rotatebox{90}{Motion Stream}}
    & Motion Stream (Walker~\etal~\cite{cmu-opticalflow-iccv2015}) &*14.3  &4.96 &21.2 &18.8  &19.0  & 15.9  \\
 	& Motion Stream (Ours-UCF)   &- &\textbf{13.9}   &51.0 &35.7  &46.4 & 32.5  \\ 
	& Motion Stream (Ours-HMDB)   &\textbf{24.1}  &-  &42.4  &30.6  &42.2 & 30.1  \\ 
	& Motion Stream (Ours-UCF+HMDB)   &-  &- &\textbf{51.1} &\textbf{35.9}  &\textbf{48.4}  & \textbf{32.7} \\ \cline{2-8}
	& $\rightarrow$ Motion Stream (Ground-truth Motion)   &38.7    &20.0   &52.4 &-  &- & -\\  \specialrule{.12em}{.1em}{.1em}
	\multirow{7}*{\rotatebox{90}{Two-Stream}}
	& Appearance + Appearance                   &     64.0       &  35.5           &  73.4 &   65.8                  &   81.3    & 65.1      \\ 
	& Appearance + Motion (Walker~\etal~\cite{cmu-opticalflow-iccv2015})  &    *64.5        &    35.9         &  73.1 &   65.9                 &     81.5      & 65.0  \\ 
	& Appearance + Motion (Ours-UCF)          &   -         &     \textbf{37.1}        &    74.5 &   67.4                &      82.1  & 66.0     \\
	& Appearance + Motion (Ours-HMDB)          &    \textbf{65.5}        &   -          &   74.3 &   67.1                &    81.9   & 65.6      \\ 
	& Appearance + Motion (Ours-UCF+HMDB)          &    -        &   -          &  \textbf{74.5}  &   \textbf{67.5}                 &    \textbf{82.3}    & \textbf{66.1}     \\   \cline{2-8} 
   & $\rightarrow$ Appearance + Motion (Ground-truth Motion)             &      68.1      &   39.5          &  77.4 &   -                 &  -   & -         \\  
   \specialrule{.12em}{.1em}{.1em}
	
    \end{tabular}
    \caption{Accuracy results (in \%) on static-image action recognition datasets. Note that for UCF/HMDB-static and PennAction, the methods train from the static center frames of the videos. Dashes indicate results that would require train/test overlaps, and hence are omitted for our approach. $\rightarrow$ indicates the performance upper bound by using ground-truth motion. *The model provided by Walker~\etal~\cite{cmu-opticalflow-iccv2015} is trained on the whole UCF-101 dataset, hence it may have some mild advantage due to overlap with the test data in the starred case. The inferred motion from our Im2Flow framework performs much better than Walker~\etal (Motion Stream---Ours vs. Walker~\etal) for recognition. Injecting our inferred motion into a standard two-stream network achieves significant gains compared to the one-stream counterpart (Two-Stream Ours vs. Appearance Stream).
    }
    \label{tab:action_accuracy}
    \vspace{-6pt}
\end{table*}

\vspace*{-0.1in}
\paragraph{Results} 
Table~\ref{table:optical_flow_eval} shows the results on UCF-101 (see Supp.~for similar results on HMDB and Weizmann). Our method outperforms both prior work and the Nearest Neighbor baseline consistently by a large margin on all datasets across all metrics. This result shows the effectiveness of the proposed motion encoding and Im2Flow network.

Fig.~\ref{fig:qualitative} shows qualitative results. Our Im2Flow network can predict motion in a variety of contexts. The structured random forest approach by Pintea~\etal~\cite{dejavu-eccv2014} makes reasonable predictions on Weizmann, but struggles on more complicated datasets such as HMDB-51. The classification approach by Walker~\etal~\cite{cmu-opticalflow-iccv2015} predicts plausible motions in many cases, but the predictions are inherently coarse and usually only depict the general trend of motion of the objects in the scene. Our Im2Flow network makes more reliable and fine-grained predictions. For example, in the baby crawling case, while~\cite{cmu-opticalflow-iccv2015} can only predict that the baby is going to move leftwards, our model predicts motion at various body parts of the baby. Similarly, in the example of a boy playing the violin, our model makes reasonable predictions across various parts of the image. Moreover, aside from human motions, our model can also predict scene motions, such as the falling waves in the ocean. However, our motion prediction model is far from perfect. It can fail especially when the motion present in the static image is subtle or the background is too diverse, as shown in the failure cases (last row) in Fig.~\ref{fig:qualitative}.

\begin{figure}
    \center
    \includegraphics[scale=0.32]{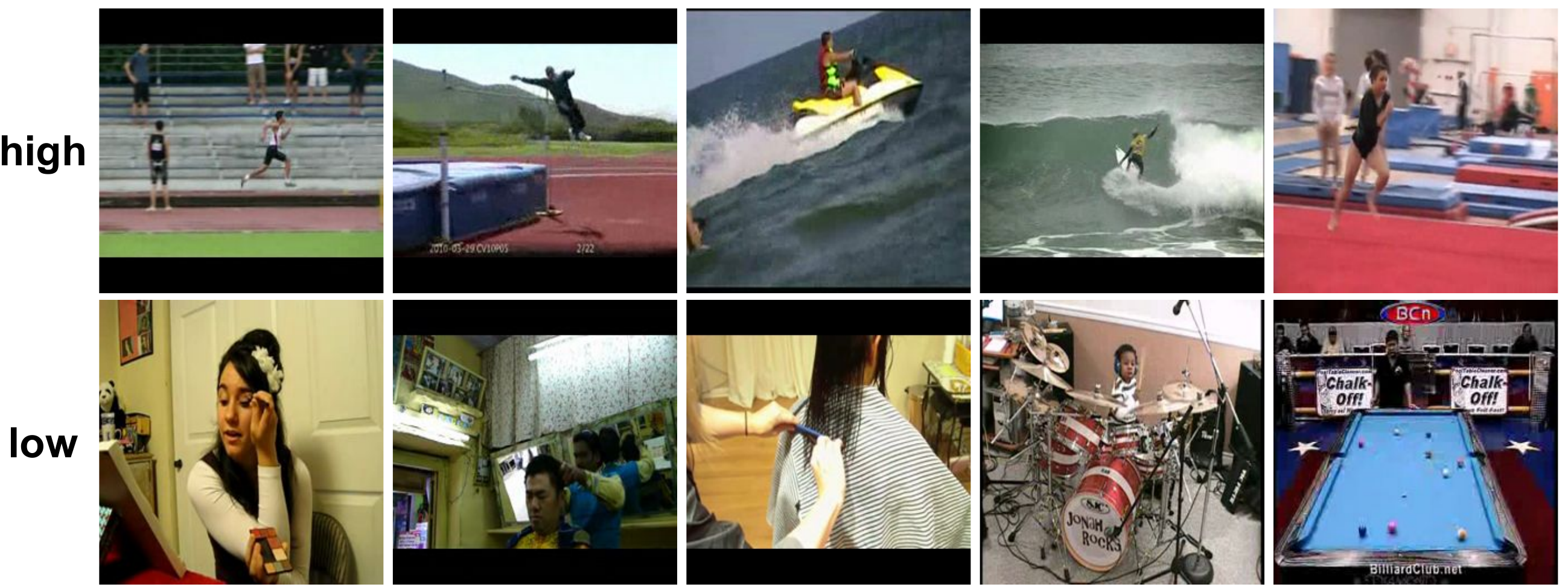}
    \caption{Examples of static images with the greatest/least motion potential determined by our Im2Flow framework.}
    \label{fig:motion_potential}
    \vspace{-2pt}
\end{figure}

With the ability to anticipate flow, Im2Flow can infer the \emph{motion potential} of a novel image---that is, the strength of movement and activity that is poised to happen. Given an image, we compute its motion potential score by inferring flow, then normalizing the average magnitude by the area of the foreground (obtained using~\cite{pixelobjectness}) to avoid a bias to large objects. Fig.~\ref{fig:motion_potential} shows static images our system rates as having the greatest/least motion potential. Motion potential offers a high-level view of a scene's activity, identifying images that are most suggestive of coming events.

\vspace*{-4pt}
\subsection{Action Recognition}
\vspace*{-4pt}

\label{sec:action_recognition}

\begin{figure}
    \vspace*{-8pt}
    \center
    \includegraphics[scale=0.51]{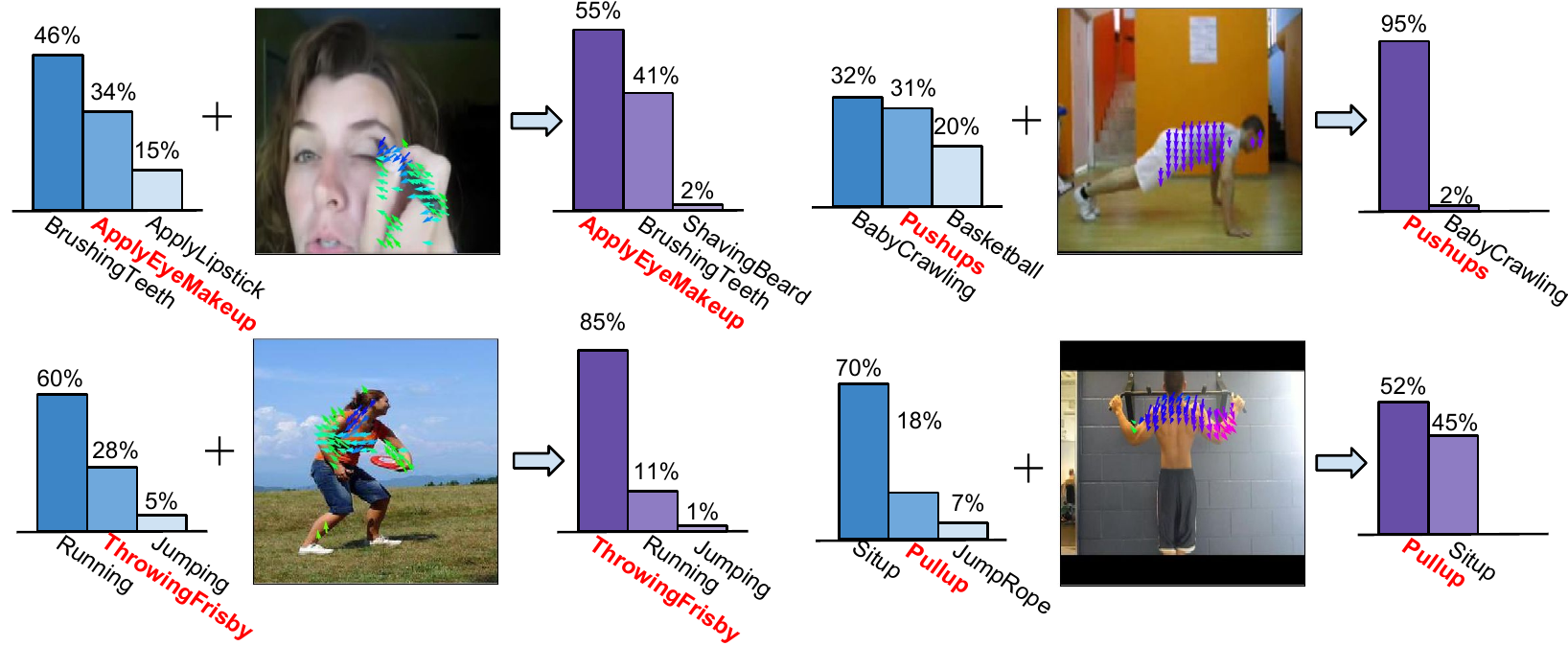}
    \caption{Examples of how the inferred motion can help static-image action recognition. For each example, the left shows the classification results of the appearance stream, and the right shows the two-stream results after incorporating the inferred motion.}
    \label{fig:action_qualitative}
    \vspace*{-8pt}
\end{figure}

Having demonstrated the accuracy of our flow estimates, we now examine the impact on static-image action recognition.
For these experiments, we use seven total datasets: three static-image datasets we construct from existing video datasets, three existing static-image benchmarks, and one dynamic scene dataset.

The three constructed datasets draw on videos from UCF-101~\cite{soomro2012ucf101}, HMDB-51~\cite{kuehne2011hmdb}, and Penn Action~\cite{zhang2013actemes}. For each, we construct static-image datasets by taking the standard train-test splits and extracting the center frame in each video.  This yields train/test sets of 9,537/3,783 (UCF-Static), 3,570/1,530 (HMDB-Static), and 1,258/1,068 (Penn Actions) images, respectively. Since they originate from video, these datasets allow us to compute the actual flow, and thereby place an upper bound on its role in static-image action recognition.

The three static-image action benchmarks are Willow~\cite{delaitre2010recognizing}, Stanford10~\cite{yao2011human,chen2013watching}, and PASCAL2012 Actions~\cite{pascal-voc-2012}.  Willow has 7 action classes, with 427 training images and 484 testing images. Stanford10 is a subset of Stanford40~\cite{yao2011human} generated in~\cite{chen2013watching}. It contains the 10 action classes most related to human action (1,000/1,672 train/test images), as opposed to being characterized by the objects that appear in the scene. For PASCAL2012 Actions, we use bounding-box-cropped images from the standard train/val sets, leading to a train/test set of 2,645/2,658 images.

In all results, we train Im2Flow with UCF-101 (Ours-UCF), HMDB-51 (Ours-HMDB), or their combination (Ours-UCF+HMDB). We then use the trained networks to predict flow images for all the static-image datasets except for the dataset the network is trained on. Thus, we test whether motion learned from disjoint videos/labels can transfer to static images from another domain.

We compare to the following baselines:
\vspace*{-0.1in}

\begin{itemize}
\itemsep0em
\item \textbf{Appearance Stream:} The recognition network is trained only on the original static images, representing the status quo in static-image action recognition.			
\item \textbf{Motion Stream (Ground-truth):} The recognition network uses ``ground-truth" optical flow computed from the video frames.  This baseline is only possible for UCF-static, HMDB-static, and PennAction.
\item \textbf{Motion Stream (Walker~\etal~\cite{cmu-opticalflow-iccv2015}):} We use the publicly available optical flow prediction model of \cite{cmu-opticalflow-iccv2015} to generate the motion stream's inputs.
\item \textbf{Appearance + Appearance:} A standard model that ensembles two separately trained appearance streams to give more robust predictions.
\end{itemize}
\vspace*{-0.1in}

We stress that all recognition baselines employ the same two-stream architecture, differing only in the source of the second stream.

Table~\ref{tab:action_accuracy} shows the action recognition results. In the top part of the table, we show the performance of using a single stream. Although the model of Walker~\etal~\cite{cmu-opticalflow-iccv2015} can predict coarse optical flow successfully in many cases as shown in Fig.~\ref{fig:qualitative}, the predicted coarse motion works poorly for recognition. The inferred motion from our Im2Flow framework performs much better, even as well as the ground-truth motion in some cases. The bottom part shows the two-stream performance after combining the appearance and motion streams. Across all six datasets, we obtain large gains (1-6\% relative gain for ours vs. appearance stream) by inferring motion. Although the test cases are from different domains than that which trained our flow network, our approach generalizes well due to the motion prior transferred from unlabeled video to static images. Additionally, we use our Im2Flow network to predict motion (independently) for each UCF-101 frame and then use the method from~\cite{wang2017temporal} to train the temporal stream using the predicted motion images. With BN-Inception as the base network, we achieve 90.5\% accuracy on UCF, comparable to the SoTA on \emph{video} and further suggesting the power of using predicted motion.

Fig.~\ref{fig:action_qualitative} shows some qualitative results from various datasets to illustrate how the inferred motion can help recognition. While a classifier solely based on appearance can be confused by actions appearing in similar contexts, the inferred motion provides cues about the fine-grained differences among these actions to help recognition. For instance, the first image shows a woman applying eye makeup. However, brushing teeth, applying eye makeup, and applying lipstick are all visually similar. Showing the hand movements of the woman guides the classifier to make the correct prediction. Moreover, our model can even make reasonable predictions for actions that do not appear in the training set, e.g., throwing frisby is a novel action in Stanford 10 dataset, but the inferred motion can still help recognition. See Supp. for more examples.

Finally, we use YUP++ Dynamic Scenes~\cite{dynamic-scene} to explore how inferred motion may benefit dynamic scene recognition. Table~\ref{Table:dynamic-scene} shows the results. We include this scenario since motion is also indicative in many dynamic natural scenes, e.g, waterfall, falling trees, rushing river. Given a static image of a dynamic scene, hallucinating motion from the scene may also help recognition. We use 90\% of the dataset (using the standard 10-90 split) to train our Im2Flow prediction network. From the remaining 10\%, we construct the ``static-YUP++" dataset for \emph{static-image} dynamic scene recognition. Specifically, we use 2/3 (from the 10\% reserved videos) as training data and 1/3 as test data. Once again, with the inferred motion, the recognition accuracy improves by a large margin (78.2\% vs.~74.3\%) compared to using only static images alone.

\begin{table}
\centering
\begin{tabular}{c?{0.5mm}c|c}
                     & Accuracy & mAP   \\ \specialrule{.12em}{.1em}{.1em}
Appearance                  & 74.3   & 79.3 \\ 
Ground-truth Motion            & 55.5   & 62.0 \\ 
Inferred Motion     & 30.0   & 37.0 \\ 
Appearance + Appearance             & 75.2   & 79.8 \\
Appearance + Inferred Motion & \textbf{78.2}   & \textbf{82.3}   \\ \hline
Appearance + Ground-truth Motion      & 79.6   & 83.6 \\ 
\specialrule{.12em}{.1em}{.1em}

\end{tabular}
\caption{Static-image dynamic scene recognition results (in \%) on the static-YUP++ dataset~\cite{dynamic-scene}. The inferred scene motion improves the recognition accuracy by a large margin.}
\label{Table:dynamic-scene}
\vspace{-6pt}
\end{table}

\vspace*{-0.1in}
\paragraph{Comparison to alternative recognition models}

The results above are all apples-to-apples, in that the only moving part is whether and how implied flow is injected into a two-stream recognition architecture.  Next we briefly  place our action recognition results on an absolute scale against reported results in the literature.

On Stanford10, the method of~\cite{chen2013watching} uses unlabeled video as a means to generate synthetic training examples in pose space for the low-shot training regime.  With 250 training images per class, their method yields 50.2 mAP, whereas our method achieves 74.9 mAP. However, note that our method also benefits from using a deep learning approach.

For Willow, Table~\ref{tab:willowresult} compares our results to several state-of-the-art methods. We attempt three variants of our approach using AlexNet, VGG-16, and ResNet-50 as the base network, respectively. Our approach combines appearance and the inferred motion, and performs well compared to all baselines. Of note, our model with VGG-16 as the base network significantly outperforms Zhang~\etal~\cite{zhang2016action}, who also use VGG-16. Without using separate body part and/or object detectors as in~\cite{liang2014expressive,mettes2016no}, our end-to-end recognition model compares favorably.

\begin{table}
\centering
\begin{tabular}{c?{0.5mm}c}
               & ~~~~~mAP (\%)~~~~~ \\ \specialrule{.12em}{.1em}{.1em}
Delaitre~\etal~\cite{delaitre2010recognizing}  & 59.6   \\ 
Sharma~\etal~\cite{sharma2013expanded}  & 67.6 \\ 
Khan~\etal~\cite{khan2014scale} & 68.0 \\ 
Zhang~\etal~\cite{zhang2016action} & 77.0  \\ 

Liang~\etal~\cite{liang2014expressive}   &   80.4  \\ 
Mettes~\etal~\cite{mettes2016no} &  81.7 \\ \hline
Ours (AlexNet as base network) &  74.0    \\ 
Ours (VGG-16 as base network)     &  87.2  \\ 
Ours (ResNet-50 as base network)  &  \textbf{90.5} \\ \specialrule{.12em}{.1em}{.1em}
\end{tabular}
\caption{Comparison to other recognition models on Willow~\cite{delaitre2010recognizing}.}
\label{tab:willowresult}
\vspace*{-6pt}
\end{table}
\vspace*{-0.1in}

\section{Conclusion}

We presented an approach to hallucinate the motion implied by a single snapshot and then use it as an auxiliary cue for static-image action recognition. Our Im2Flow framework achieves state-of-the-art performance on optical flow prediction from an individual image. Moreover, using a standard two-stream network, it enhances recognition of actions and dynamic scenes by a good margin. As future work, we plan to explore hierarchical representations to encode the temporal evolution of multiple video frames.

\noindent\textbf{Acknowledgements:} 
This research was supported in part by an ONR PECASE Award N00014-15-1-2291 and an IBM Faculty Award and IBM Open Collaboration Award. We thank Suyog Jain, Chao-Yeh Chen, Aron Yu, Yu-Chuan Su, Tushar Nagarajan and Zhengpei Yang for helpful input on experiments or reading paper drafts, and also gratefully acknowledge a GPU donation from Facebook.

\newpage
{\small
\bibliographystyle{ieee}
\bibliography{cv_archive,kg-refs-flow}
}

\end{document}